\documentclass[letterpaper, 10 pt, conference]{ieeeconf}  
\usepackage[T1]{fontenc}
\usepackage{algorithm}
\usepackage{xcolor}
\usepackage{algpseudocode}
\usepackage{ upgreek }
\usepackage{multicol}
\usepackage{ dsfont,setspace }
\usepackage{amsmath,graphicx}
\usepackage{amssymb}
\usepackage{amsmath}

\usepackage{lineno}
\usepackage{epstopdf}
\usepackage{authblk}
\usepackage{dblfloatfix}
\usepackage{tabularx, makecell}
\newcolumntype{L}{>{\raggedright\arraybackslash}X}
\usepackage[
  separate-uncertainty = true,
  multi-part-units = repeat
]{siunitx}

\usepackage{subcaption}
\usepackage[colorlinks]{hyperref}
\hypersetup{
     colorlinks,
     linkcolor=black,
     citecolor=black,
     filecolor=black,
     urlcolor=blue,
 }
\usepackage{cleveref}
\usepackage{textcomp}
\usepackage{gensymb}
\usepackage{algorithm}
\usepackage{algpseudocode}

\IEEEoverridecommandlockouts                              

\begin{document}

\title{\LARGE \bf
Reducing Onboard Processing Time for Path Planning in Dynamically Evolving Polygonal Maps}

\author{
Aditya Shirwatkar$^{1}$, Aman Singh$^{1}$, Jana Ravi Kiran$^{2}$
\thanks{This work is done as a project for the course CP:230 Motion Planning and Autonomous Navigation}
\thanks{$^{1}$A. Shirwatkar (email: \href{mailto:adityasr@iisc.ac.in}{adityasr@iisc.ac.in}), A. Singh (email: \href{mailto:saman@iisc.ac.in}{saman@iisc.ac.in}), are with the Robert Bosch Center for Cyber Physical Systems, Indian Institute of Science, Bengaluru.}
\thanks{$^{2}$J. Kiran (email: \href{mailto:ravikj@iisc.ac.in}{ravikj@iisc.ac.in}) is with the Department of Aerospace Engineering, Indian Institute of Science, Bengaluru.}
}%

\maketitle
\thispagestyle{empty}
\pagestyle{empty}

\begin{abstract}

Autonomous agents face the challenge of coordinating multiple tasks (perception, motion planning, controller) which are computationally expensive on a single onboard computer. To utilize the onboard processing capacity optimally, it is imperative to arrive at computationally efficient algorithms for global path planning. In this work, it is attempted to reduce the processing time for global path planning in dynamically evolving polygonal maps. In dynamic environments, maps may not remain valid for long. Hence it is of utmost importance to obtain the shortest path quickly in an ever-changing environment. To address this, an existing rapid path-finding algorithm, the Minimal Construct was used. This algorithm discovers only a necessary portion of the Visibility Graph around obstacles and computes collision tests only for lines that seem heuristically promising. Simulations show that this algorithm finds shortest paths faster than traditional grid-based A* searches in most cases, resulting in smoother and shorter paths even in dynamic environments.
\end{abstract}

\textbf{Keywords:} \textit{Path Planning, Navigation, Mobile Robots}

\section{Introduction}

Autonomous agents, such as drones and self-driving cars, have revolutionized many industries, from transportation to agriculture. However, these agents face the challenge of coordinating multiple tasks on a single onboard computer. These tasks include perception, motion planning, and control, which can be accomplished using software components available in the ROS framework \cite{1}.
But all of these tasks are are computationally expensive, hence it is critical for mobile robots which operate in dynamic and unpredictable environments to require quick and accurate decision-making capabilities.

One of these essential tasks for autonomous agents is global path planning, which involves finding an optimal path from a start location to a goal location while avoiding obstacles. Traditional path planning methods rely on a rectangular grid-based occupancy map, where each cell represents a small spatial unit that can be blocked, free, or unknown. The map is then searched cell-by-cell to find the shortest path using different versions of A* or Dijkstra's algorithms. These can be inefficient for large-scale environments. Moreover, these algorithms require discretization of the environment, leading to sub-optimal paths and limited robot motion. Despite this, most of the current state-of-the-art navigation software still uses these methods \cite{2}.

\begin{figure}
    \centering
    \includegraphics[ width=\linewidth]{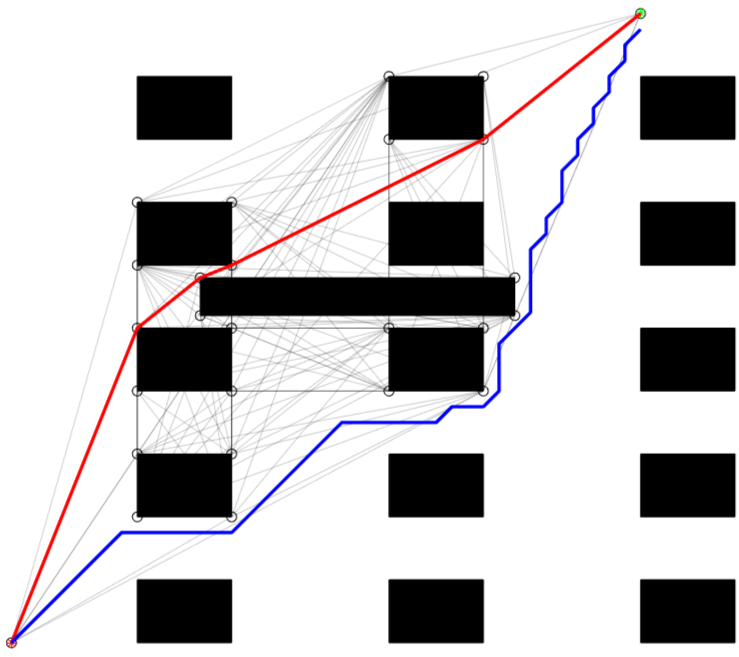}
    \captionsetup{font=small}
    \caption{The figure shows the obstacle-avoiding shortest paths in a polygonal map. The black polygonal areas represent the walls expanded by the robot's size. The minimal construct algorithm explores only a small fraction of the visibility graph of the map, as shown by the thin grey lines. In comparison, the jagged path found by A* search (shown in blue) in an equivalent grid is sub-optimal in length. On the other hand, the visibility graph-based minimal construct algorithm (shown in red) provides an optimal path with minimal complexity of only four line segments.}
    \label{fig:main_example_intro}
\end{figure}

Polygonal maps offer several advantages over grids, which make them a promising alternative for mobile robot navigation in dynamic environments. Unlike grids, geometric arrangements have a compact memory footprint and are well-suited to represent moving obstacles. Additionally, they are not susceptible to the problems of discretization. Most significantly, polygonal maps give rise to the Visibility Graph \cite{3}, which comprises edges of the polygons in the scene and additional edges that connect pairs of polygon corners that can "see" each other. The shortest paths can be found in this graph, and once it is constructed, the A* algorithm can quickly find smooth paths of optimal length with a minimal number of line segments. This is in stark contrast to the jagged and sub-optimal paths that emerge from a grid-based approach. A visual representation is provided in Figure. \ref{fig:main_example_intro}

Although the search itself is relatively fast, constructing the Visibility Graph requires significantly more computation time. The most efficient algorithms known today require $O(n^2)$ operations, where $n$ is the number of polygon edges. Therefore, the time required to construct the graph is highly dependent on the complexity of the map. Additionally, the frequent changes in the environment, such as moving objects, people, or exploration of uncharted areas, can make the graph obsolete. In such cases, the graph needs to be reconstructed or repaired, which slows down the process of finding the shortest path.

\begin{figure}[htp!]
    \centering
    \includegraphics[width=0.8\linewidth]
    {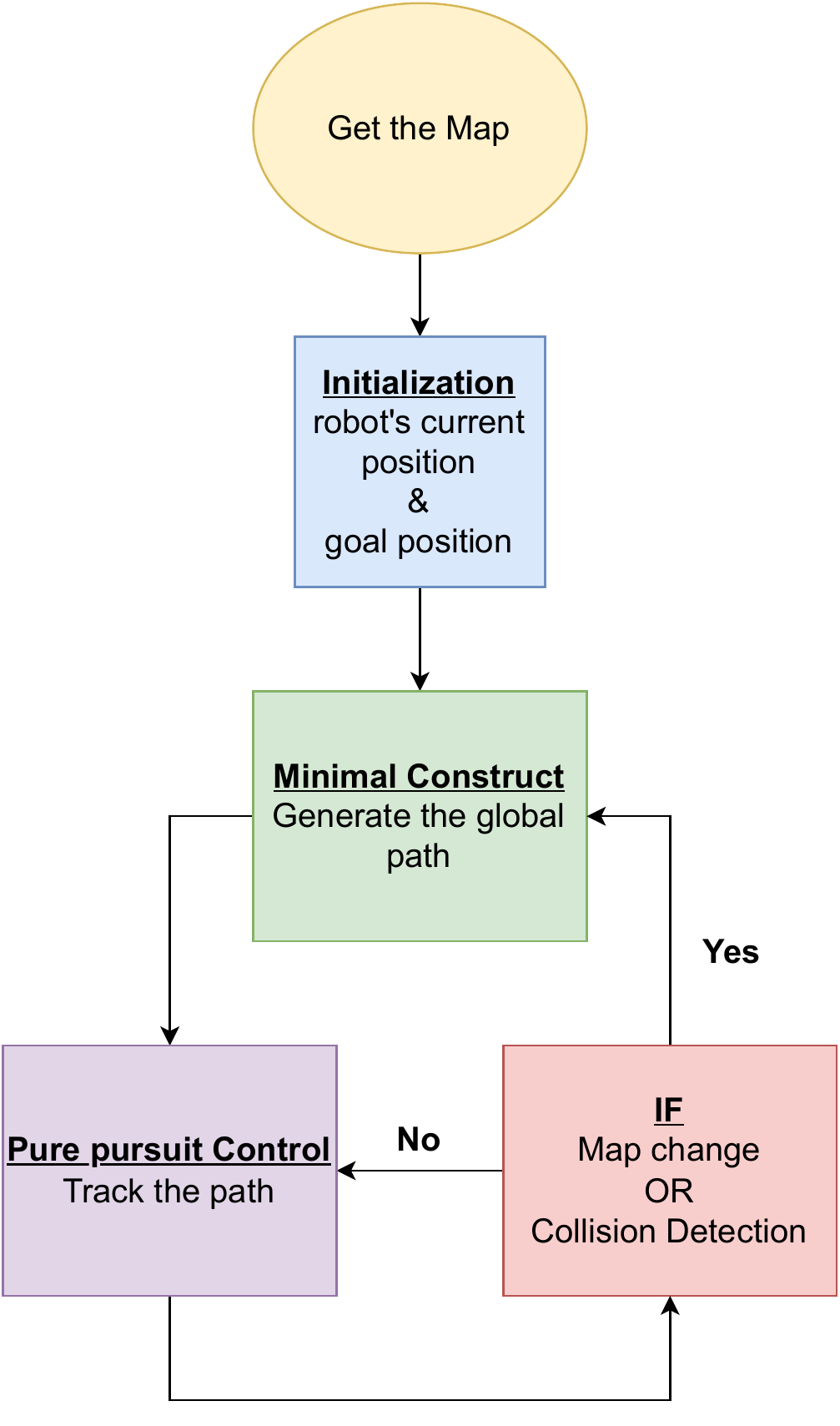}
    \caption{Overview of the approach}
    \label{Overview}
\end{figure}

The Minimal Construct algorithm \cite{3} addresses this problem by only computing the required portion of the Visibility Graph during an A* search, rather than pre-computing it. The algorithm always begins with the smallest possible graph, which is a straight line from the start to the destination. In case this line intersects with an obstacle, only then it goes on to update the graph. This approach reduces the computational expense of the algorithm by limiting the number of line intersection tests to only those lines that appear promising based on heuristics. The features described above make the Minimal Construct algorithm one of the best choices to reduce the computation time for path planning in dynamically evolving polygonal maps. 


Thus, the goal of this work is to propose a pipeline that employs a fast global path planner to reduce the time required for map re-computation as the map changes. We do this because shorter computation durations reduce the load on a single onboard CPU while simultaneously shortening reaction times. We use Minimal Construct as a global planner in dynamic environments.
To track the given global plan, we use the Pure Pursuit Controller \cite{4} as the local planner. And we perform re-computation of the global path, only when there is a change in the environment (eg. obstacle disappeared, appeared, or moved) or the current path collides with any obstacle. The overview of the approach is given in Fig \ref{Overview}. We compare results of or method with the Gird-based A* algorithm for the path planning, in one static environment case and four dynamically evolving environment cases. We found out that the Grid-based A* algorithm give a sub-optimal path from the source to the target. On the other hand, the proposed method gives an optimal route. We also found that the map re-computation time was significantly lesser in the proposed method as compared to the Grid-based A*.


\section{Methodology}

\subsection{Global Planner}

\subsubsection{Preliminaries}

The Visibility Graph can be built using an $O(n^3)$ algorithm, which examines all $O(n^2)$ pairs of vertices and checks if the edges between them cross with any of the $n$ lines in the picture. However, this method necessitates a large number of line intersection computations.
\begin{figure}
    \centering
    \includegraphics[width=\linewidth]{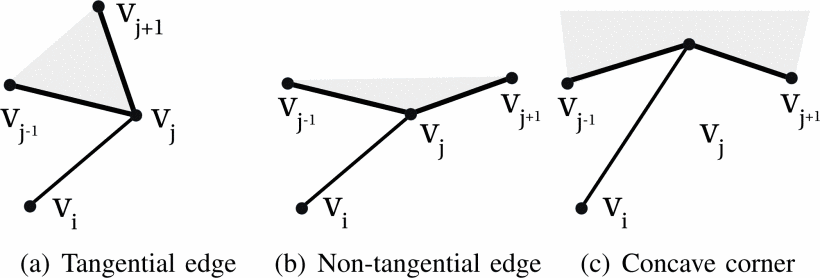}
    \captionsetup{font=small}
    \caption{a) An edge $(v_i, v_j)$ is considered tangential at vertex $v_j$ if the adjacent corners of the polygon, $v_{j-1}$ and $v_{j+1}$, are positioned on the same side of the edge. If the edge $(v_i, v_j)$ is not tangential (b), or if vertex $v_j$ is concave (c), then the triangle $\delta(v_i, v_{j-1}, v_{j+1})$ is entirely visible from vertex vi. Consequently, the edge $(v_i, v_j)$ cannot be included in the shortest path.}
    \label{fig:edge_example}
\end{figure}

\begin{algorithm}
\caption{Minimal Construct}
\label{algo:1}
\small
\textbf{Input:} {Set of polygons S, Start vertex s, Target vertex t}

\textbf{Output:} {Path P}
\begin{algorithmic}[1]
\State Priority $\mathbf{q}$ \Comment{Initialize Priority Queue}
\State Graph G = (V , E) \Comment{Initialize an Empty Graph}
\State $V \leftarrow V \cup {s, t}$ \Comment{Add Source($s$) and Target Node($t$)}
\State $E \leftarrow E \cup {(s, t)}$ \Comment{Add $t$ and $s$ as a neighbours of each other} 
\State SETPARENT(s, t) \Comment{set $s$ as a parent of $t$}
\State CLOSE(s) \Comment{Mark $s$ as close}
\State PUSH(t, q) \Comment{Push $t$ in the priority queue i.e. Mark it open}
\While{q is not empty}
    \State $v \leftarrow POP(q)$ \Comment{Pop the vertex $v$ with lowest priority ($f$)}
    \State $u \leftarrow PARENT OF(v)$ \Comment {Find the parent of $v$}
    \State Polygon p $\leftarrow$ LINEINTERSECTIONTEST(S,(u, v))
    \If{(p == nil)} \Comment{If no Intersection then}
        \If{(v = t)} \Comment{If target reached}
            \State $P \leftarrow EXTRACTPATH(v)$ \Comment{Backtrack parents}
            \State \Return P \Comment{\textbf{Finished!}}
        \EndIf
        \State CLOSE(v) \Comment{close $v$}
        \ForAll{$(vi, (vi, v) \in E)$} \Comment{For all neighbours of $v$}
            \If{(!ISCLOSED(vi))} \Comment{If $v_i$ is closed}
                \If{(!ISOPEN(vi) or g(v) + $|v - vi|$ $<$ g(vi))}
                    \State SETPARENT$(v, vi)$ \Comment{set $v_i$ as parent of $v$}
                    \State $g(vi) \leftarrow g(v) + |v - vi|$ \Comment{Update cost $g$}
                    \State $h(vi) \leftarrow |vi - t|$ \Comment{Update heuristics $h$}
                    \State $f(vi) \leftarrow g(vi) + h(vi)$ \Comment{Update priority $f$}
                    \State PUSH(vi, \textbf{q}) \Comment{Push $v_i$ in priority queue} 
                \EndIf
            \EndIf
        \EndFor
    \Else \Comment{If polygon $\mathbf{p}$ has intersected}
        \State $E \leftarrow E \setminus (v, u)$ \Comment{$v$ no longer parent of $u$}
        \State REMOVEPARENT(v) \Comment{Remove parent of $v$}
        \State FINDPARENT(v) \Comment{Algorithm 2}
        \If{(!ISCLOSED(p))} \Comment{If polygon $\mathbf{p}$ is not closed}
            \State CONNECTOBSTACLE(p) \Comment{Algorithm 3}
            \State CLOSE(p) \Comment{Close the intersected polygon}
        \EndIf
    \EndIf
\EndWhile
\State \Return P \Comment{Search failed. Return empty path!}
\end{algorithmic}
\end{algorithm}

In contrast, the Minimal Construct technique does not seek to compute the whole Visibility Graph. Rather, it only explores a small piece of the graph. To avoid running collision checks on line segments that do not contribute to the shortest path, the algorithm delays computing line intersection tests until they are required. 

The algorithm also makes use of the fact that concave corners and edges that are not tangential at both ends cannot be included in the shortest path. An edge $(v_i,v_j)$ is considered tangential at vertex $v_j$ if both adjacent polygon corners, $v_{j-1}$ and $v_{j+1}$, lie on the same side of the edge $(v_i, v_j)$. A visual example of this is shown in Figure \ref{fig:edge_example}. Whenever concave corners or non-tangential edges are discovered, they are immediately discarded.

\subsubsection{Minimal Construct Algorithm}

\begin{figure*}[h!]
    \centering
    \includegraphics[width=0.8\textwidth]
    {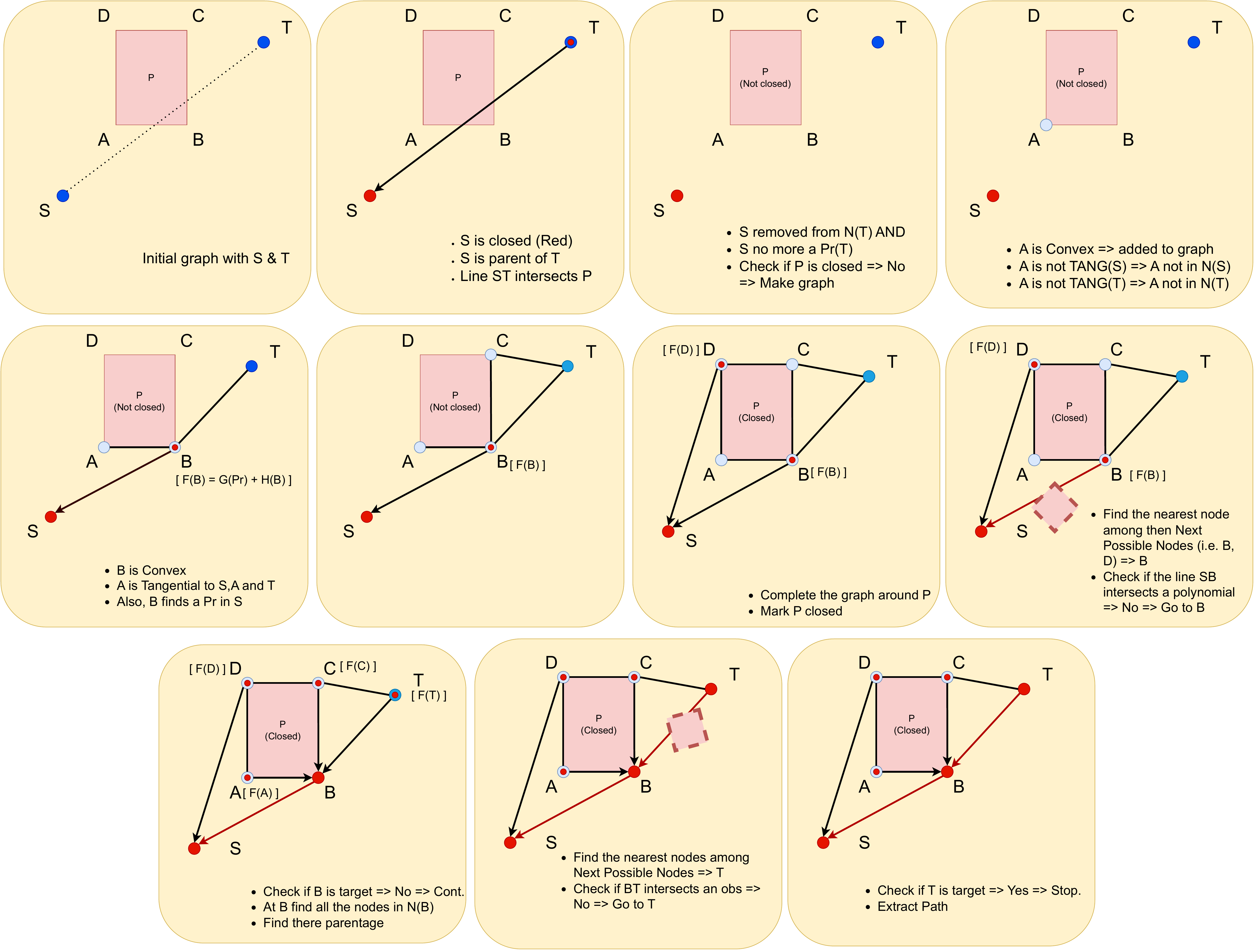}\hfill
    \caption{An example to showcase the working of Minimal Construct Algorithm}
    \label{Minimal_Construct_Ex}
\end{figure*}

Consider a scenario where the environment consists of a set of non-convex polygons which are disjoint and intersect only at endpoints. Let's represent this set of polygons with $\mathbf{S}$. The minimal construct algorithm discovers a graph $\mathbf{G}=(\mathbf{V}, \mathbf{E})$ during its search, where $\mathbf{V}$ is the set of nodes (points) in the graph and $\mathbf{E}$ is the set of edges (line segment) connecting these nodes. The set of nodes in $\mathbf{G}$ i.e. set $\mathbf{V}$ is a subset of the vertices of the polygons in the $\mathbf{S}$. The Minimal construct algorithm connects pairs of vertices in $\mathbf{V}$ with edges $\mathbf{E}$. Please note the start node $s$, and the target node, $t$, are considered as two additional points in the graph.

All the nodes connected to a node $P$ through an edge are said to be in $neighbourhood$ of $P$. As the algorithm explores the graph, all the nodes that it has explored are called $closed$ $nodes$ and all the nodes that can be explored in the next iteration step are called $open$ $nodes$. Please note that for a node to be open it should have at least one closed node in its neighbourhood. The closest, closed node in the neighborhood of a node $P$ is said to be the $parent$ of the node $P$.

All the vertices of the polygons which form an external angle greater than 180° are called $Convex$ $vertices$. An edge $(v_i, v_j)$ is considered $tangential$ at vertex $v_j$ if the adjacent corners of the polygon, $v_{j-1}$ and $v_{j+1}$, are positioned on the same side of the edge.

Minimal Construct algorithm uses A* algorithm as the basis. In order to use it, the minimal construct algorithm defines a parent of a node, as described previously. After the search is finished, we use the unidirectional parent relationship to extract the shortest path by following the path from the target parent to the parent to the start. Algorithm \ref{algo:1} uses the functions $\text{PARENTOF}(v_i)$, $\text{SETPARENT}(v_i, v_j)$, and $\text{REMOVEPARENT}(v_i)$ to set and remove parent associations between vertices. $\text{SETPARENT}(v_i, v_j)$ makes vertex $v_i$ as the parent of $v_j$, whereas $\text{REMOVEPARENT}(v_i)$ cancels the parent of $v_i$.

The A* algorithm employs the functions $\text{ISOPEN}(v_i)$, $\text{CLOSE}(v_i)$, and $\text{ISCLOSED}(v_i)$ to keep track of the vertices that have been opened or closed during the search. Vertices are opened when they are pushed into the priority queue and closed when they are removed. During the search, we can use these functions to query and manipulate the open and closed states of vertices.

Here, the pseudo-code of the Minimal Construct Algorithm, as shown in Algorithm  \ref{algo:1} is explained. To start the search, the Minimal construct algorithm forms a graph with only two nodes, the start and the target node. It connects the two nodes with a straight edge, making them each other's neighbours. It also marks the start node as closed and sets the start node as the parent of the target node. It then selects the target node as an open node by sending it to the priority queue. Please note, that the algorithm sets the start node as 
closed in the starting itself, as the start node is considered explored in the start, and it remains closed throughout. After this, the algorithm enters a loop where in each iteration it removes the vertex $v$ with lowest priority from the priority queue. Here priority represents the heuristic of the vertex, i.e. the sum of the distance from the start node, as travelled through the graph, and the straight line distance to the target node. After removing the vertex it finds its parent, $u$ and the edge $(u,v)$ is tested for Line Intersection test.
It is a computationally intensive task, and is discussed separately in following section.

If the line intersection test on the edge $(u,v)$ resulted in no intersection with polygon, the Algorithm goes ahead with the A* expansion. The algorithm checks if the popped vertex $v$ is the target vertex. If $v$ is the target vertex, then the algorithm extracts the path by following the parent pointers back to the start vertex, and returns the path as the output of the algorithm. If $v$ is not the target vertex, it closes it, meaning that it will not be expanded again during the search. It then expand the neighbors of the popped vertex $v$. For each neighbor that has not yet been closed, the algorithm checks if it is already in the priority queue. If the neighbor is not in the priority queue, or if its estimated path cost from the start vertex is lower than its current estimated path cost, the algorithm updates its parent, path cost, and priority, and adds it to the priority queue. It updates the path cost $g$ of the neighbor as the sum of the path cost of the popped vertex and the distance between the two vertices. It then calculates the heuristic estimate $h$ of the distance from the neighbor to the target vertex, and finally calculates the priority $f$ of the neighbor as the sum of its path cost $g$ and its heuristic estimate $h$. It then pushes the neighbor into the priority queue.

\begin{algorithm}[H]
\caption{Find Parent}
\label{algo:2}
\small
\textbf{Input:} {Vertex $v$}
\begin{algorithmic}[1]
\State $minPathCost \leftarrow \infty$ \Comment{Init the min path cost with infinity}
\State $u \leftarrow \text{nil}$ \Comment{Init the new parent with nil}
\ForAll{neighbors $(vi, (vi, v) \in E)$} \Comment{For all neighbors of $v$}
	\If{ISCLOSED($vi$)}
		\If{($g(vi) + |vi - v| < minPathCost$)}
			\State $minPathCost \leftarrow g(vi) + |v - vi|$ 
                \State \Comment{Updtae min cost}
			\State $u \leftarrow vi$ \Comment{Remember $v_i$ as new parent}
		\EndIf
	\EndIf
\EndFor
\If{($u \neq \text{nil}$)}
	\State SETPARENT($u$, $v$) \Comment{Set $u$ as new parent of $v$}
	\State $g(v) \leftarrow g(u) + |v - u|$ \Comment{Update path cost $g$}
	\State $f(v) \leftarrow g(v) + h(v)$ \Comment{Update priority $f$}
	\State PUSH($v$, $q$) \Comment{Push $v$ into the priority queue}
\EndIf
\Return
\end{algorithmic}
\end{algorithm}

\begin{algorithm}[H]
\caption{Connect Obstacle}
\label{algo:3}
\small
\textbf{Polygon $p$}
\begin{algorithmic}[1]
\ForAll{$v_i$ in $p$} \Comment{For all corners $v_i$ of $\mathbf{p}$}
    \If{isConvex($v_i$)} \Comment{If the corner $v_i$ is convex}
        \State $V\leftarrow V\cup{v_i}$ \Comment{Add the corner $v_i$ to the graph}
        \ForAll{$v_j$ in $V$, $j\neq i$} \Comment{For all known vertices $v_j$}
            \If{isTangential($v_i$, $v_j$)}
                \State $E\leftarrow E\cup{(v_i, v_j)}$ \Comment{Make $v_i$ a neighbor of $v_j$}
            \EndIf
        \EndFor
        \State FINDPARENT($v_i$)
    \EndIf
\EndFor
\Return
\end{algorithmic}
\end{algorithm}

If the line intersection test on the edge between the parent $u$ and popped vertex $v$ returns intersection, meaning that the edge intersects a polygon, the algorithm proceeds to handle the intersected polygon $\mathbf{p}$. First, it removes the edge between $v$ and its parent $u$ from the graph by deleting the corresponding entry from the set of edges $\mathbf{E}$. Then, it removes the parent of $v$ by setting it to nil. The algorithm then calls the $\text{FINDPARENT}(v)$ function, as shown in Algorithm \ref{algo:2}, to find a new parent for $v$.If such a parent is discovered, $v$'s priority $f(v)$ and path cost $g(v)$ are modified, and it is reinserted into the queue. This re-parenting and reinsertion of $v$ ensures that the A* search remains complete and returns the graph to the condition it would have been in if the invalid edge $(u, v)$ had not been there.  It is important to note that $v$ may remain without a parent until it is revisited at a later stage.

For each corner $v_i$ in the set of vertices of the polygon p, the algorithm checks if the corner is convex or not. If the corner $v_i$ is convex, the algorithm adds it to the set of vertices $\mathbf{V}$. For each known vertex $v_j$ in the set $\mathbf{V}$, the algorithm checks if $v_i$ and $v_j$ are tangential using the $\textbf{isTangential}(vi, vj)$ function. If $v_i$ and $v_j$ are tangential, the algorithm adds an edge $(vi, vj)$ to the set $\mathbf{E}$ of edges in the graph $\mathbf{G}$, indicating that the corners $v_i$ and $v_j$ are connected. The algorithm then calls the $\textbf{FINDPARENT}(v_i)$ function for the corner vi, which searches for the parent of $v_i$. Finally, the algorithm repeats this process for each corner of the polygon until all corners have been processed. The algorithm then returns the updated graph $\mathbf{G}$. The polygon is then closed, so that it is not added to graph multiple times. The process is repeated until either a solution is identified or the entire graph has been investigated and no solution is found.

\subsubsection{Line Intersection Test}

The Minimal Construct Algorithm uses a Line Intersection test function to determine whether an edge between the popped vertex $v$ and its parent $u$ is intersecting any polygon or not. If it intersects, the algorithm makes the graph around the intersecting polygon. Checking line intersection is a computationally intensive task, and its algorithm is discussed as follows. 

The Line intersection algorithm finds all intersection points of the line segment with the polygon. It then divides the line segment into sub-line segments using the intersection points. For each sub-line segment, it checks if the midpoint of the sub-line segment lies inside the polygon. If the midpoint is inside the polygon, then the line is intersecting the polygon. The algorithm stops there. If the midpoint is not inside the polygon, algorithm moves to the next sub-line segment. It repeats these steps for all sub-line segments. If none of the midpoints are inside the polygon, then the line segment does not intersect the polygon.

\begin{figure*}[htp!]
    \centering
    \begin{subfigure}{0.9\linewidth}
        \centering
        \includegraphics[trim={2.4cm 1cm 2cm 1.3cm},clip, width=0.24\linewidth]
        {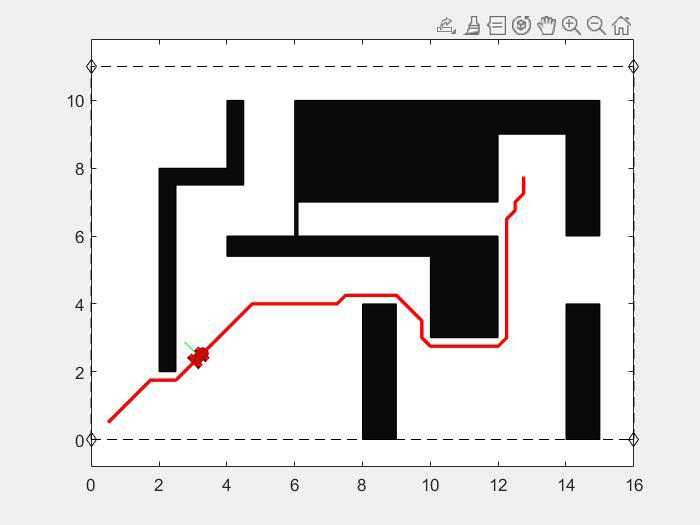}\hfill
        \includegraphics[trim={2.4cm 1cm 2cm 1.3cm},clip, width=0.24\linewidth]
        {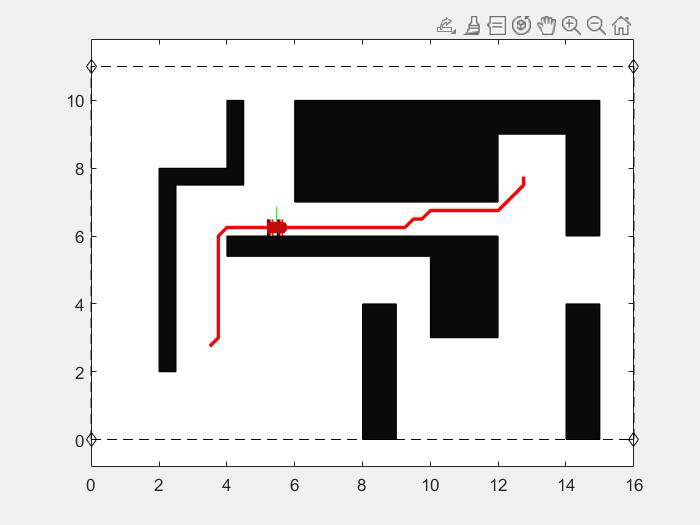}\quad
        \includegraphics[trim={2.4cm 1cm 2cm 1.3cm},clip, width=0.24\linewidth]
        {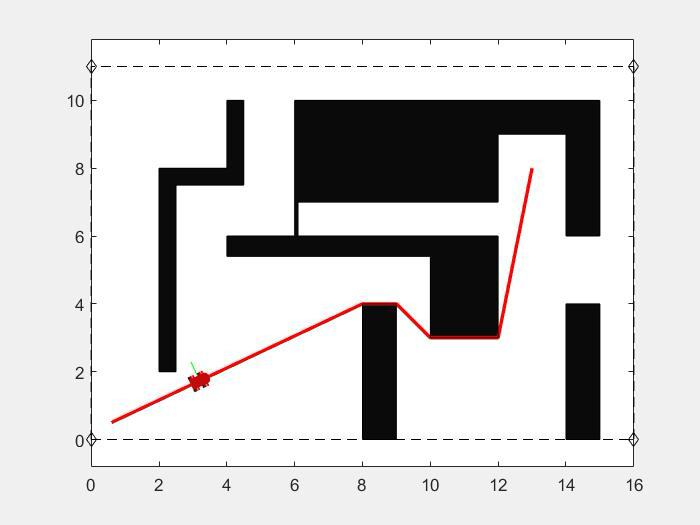}\hfill
        \includegraphics[trim={2.4cm 1cm 2cm 1.3cm},clip, width=0.24\linewidth]
        {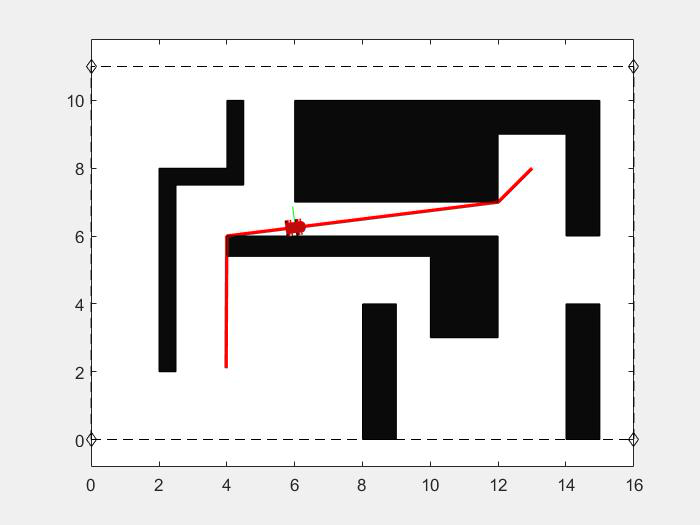}\hfill
        \caption{Case 1: One obstacle disappears in the middle of the environment}
    \end{subfigure}\quad
    \vspace{1em}
    \begin{subfigure}{0.9\linewidth}
        \centering
        \includegraphics[trim={2.4cm 1cm 2cm 1.3cm},clip, width=0.24\linewidth]
        {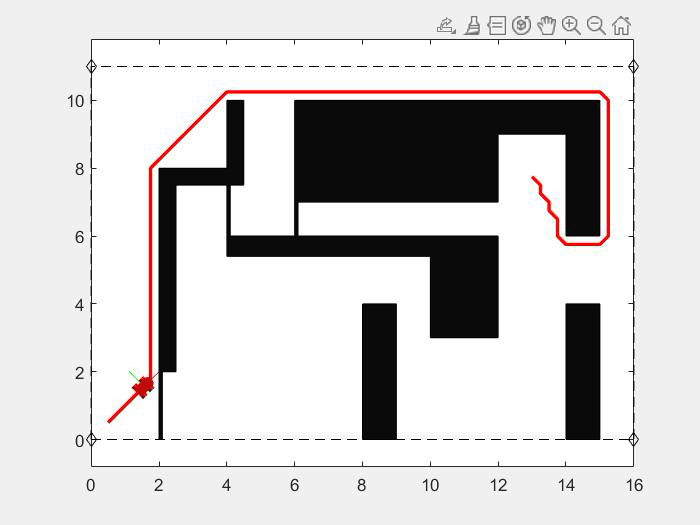}\hfill
        \includegraphics[trim={2.4cm 1cm 2cm 1.3cm},clip, width=0.24\linewidth]
        {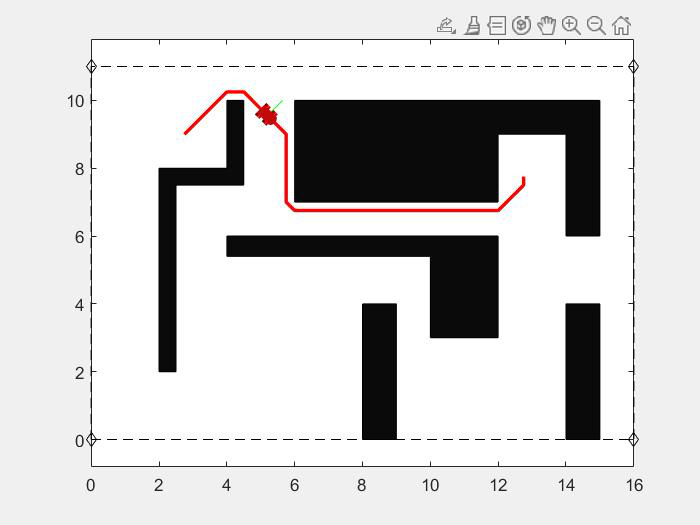}\quad
        \includegraphics[trim={2.4cm 1cm 2cm 1.3cm},clip, width=0.24\linewidth]
        {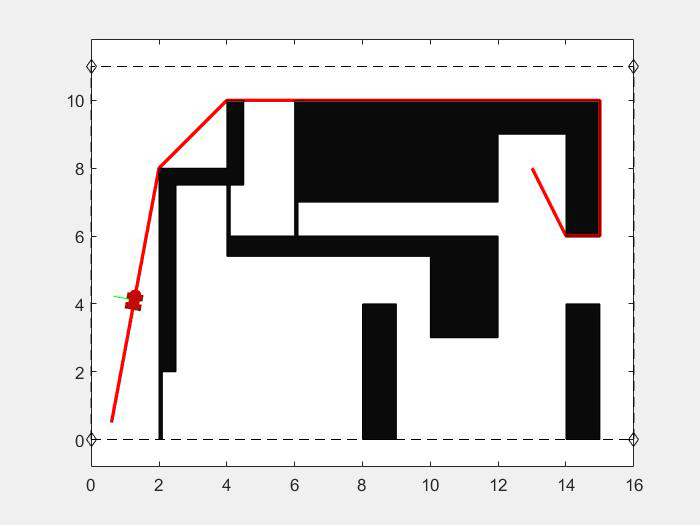}\hfill
        \includegraphics[trim={2.4cm 1cm 2cm 1.3cm},clip, width=0.24\linewidth]
        {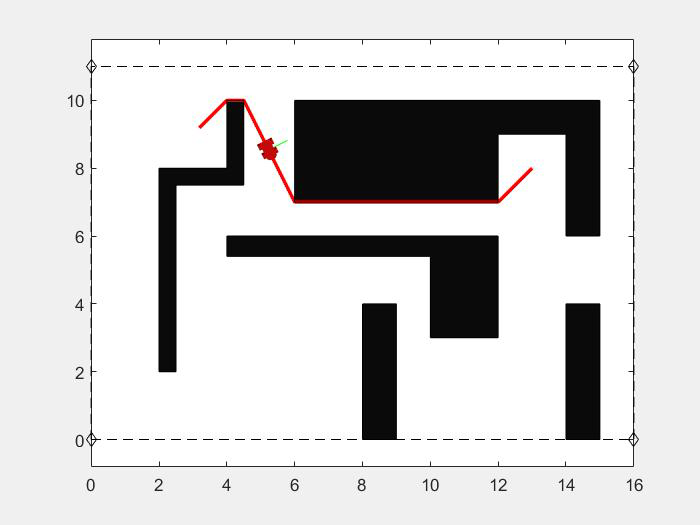}\hfill
        \caption{Case 2: Three obstacles disappear from the environment}
    \end{subfigure}\quad
    \vspace{1em}
    \begin{subfigure}{0.9\linewidth}
        \centering
        \includegraphics[trim={2.4cm 1cm 2cm 1.3cm},clip, width=0.24\linewidth]
        {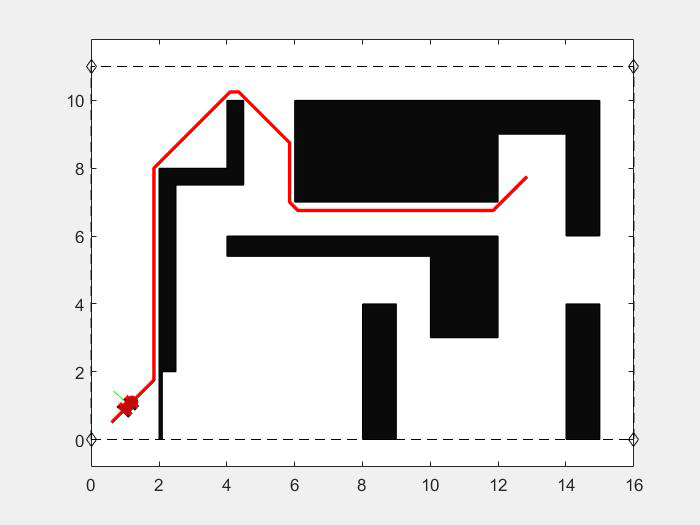}\hfill
        \includegraphics[trim={2.4cm 1cm 2cm 1.3cm},clip, width=0.24\linewidth]
        {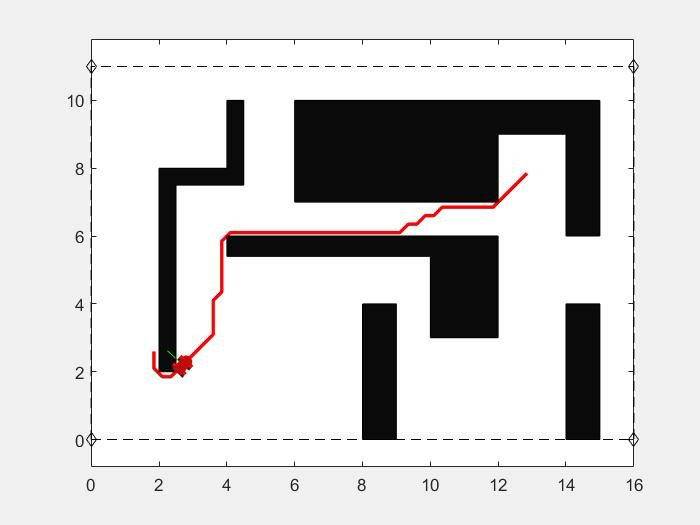}\quad
        \includegraphics[trim={2.4cm 1cm 2cm 1.3cm},clip, width=0.24\linewidth]
        {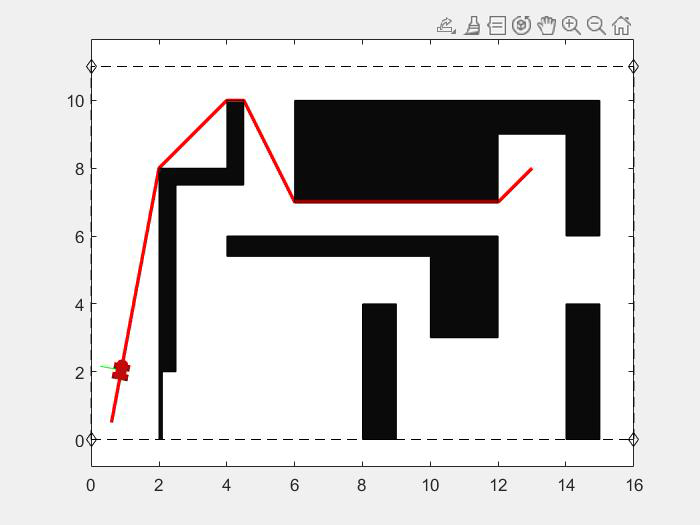}\hfill
        \includegraphics[trim={2.4cm 1cm 2cm 1.3cm},clip, width=0.24\linewidth]
        {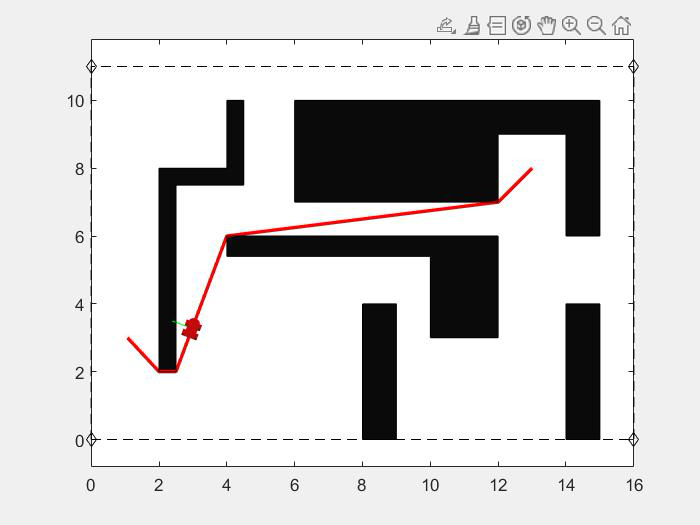}\hfill
        \caption{Case 3: One obstacle near the start location disappears from the environment}
    \end{subfigure}\quad
    \vspace{1em}
    \begin{subfigure}{0.9\linewidth}
        \centering
        \includegraphics[trim={2.4cm 1cm 2cm 1.3cm},clip, width=0.24\linewidth]
        {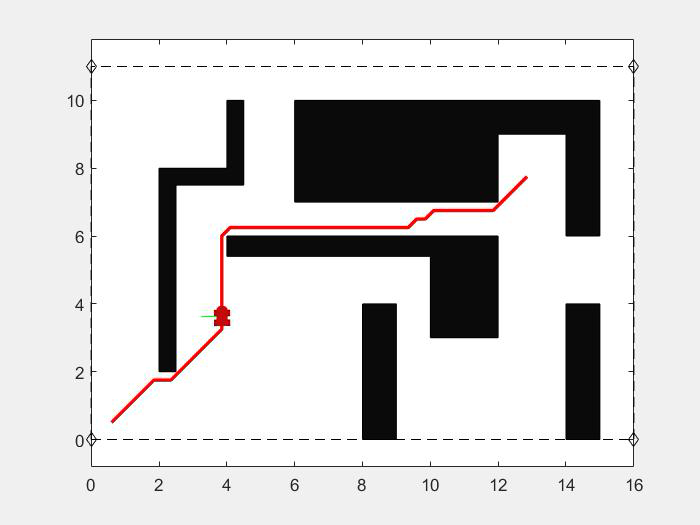}\hfill
        \includegraphics[trim={2.4cm 1cm 2cm 1.3cm},clip, width=0.24\linewidth]
        {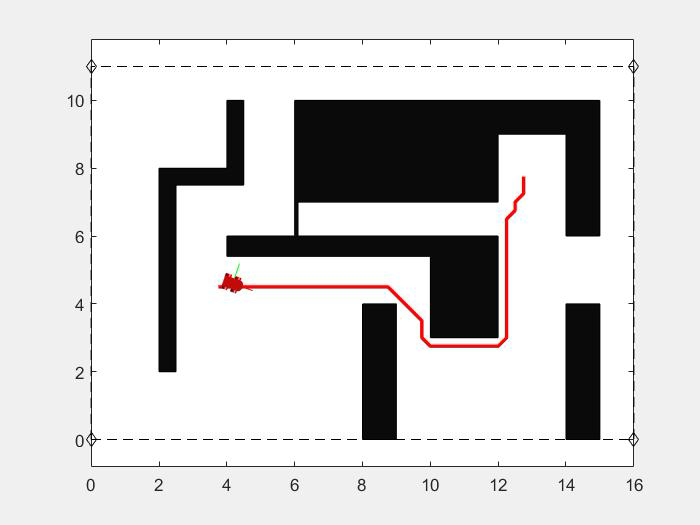}\quad
        \includegraphics[trim={2.4cm 1cm 2cm 1.3cm},clip, width=0.24\linewidth]
        {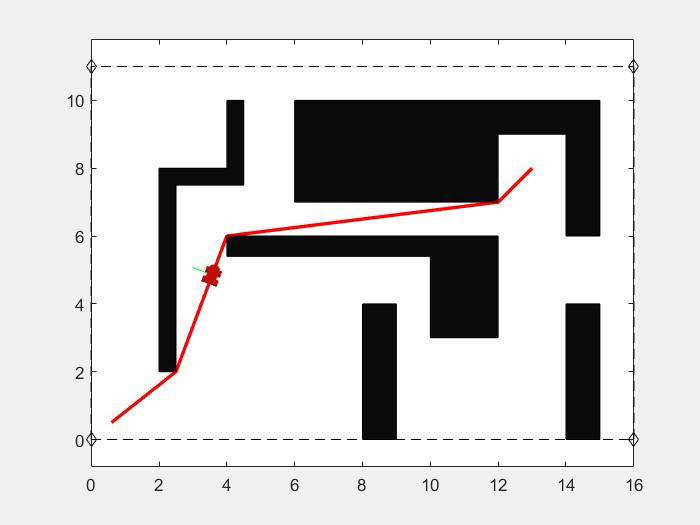}\hfill
        \includegraphics[trim={2.4cm 1cm 2cm 1.3cm},clip, width=0.24\linewidth]
        {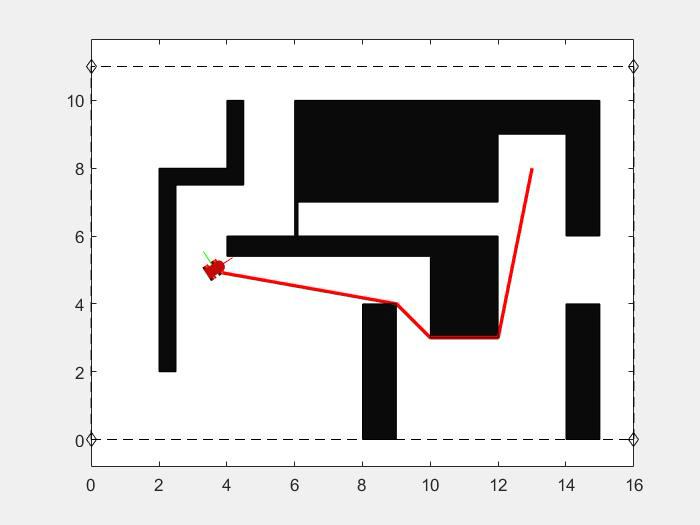}\hfill
        \caption{Case 4: One obstacle appears in the middle of the environment}
    \end{subfigure}

    \caption{In each subfigure, the two leftmost figures show the global path with grid-based A* algorithm before and after the map changed. Similarly, the two rightmost two figures show the global path with the Minimal Construct algorithm before and after the change in the map}
    \label{fig:results2}
\end{figure*}

\subsection{Local Planner}

After generating the global plan using the Minimal Construct algorithm, the next step is to track the plan using a local planner. For this task, Pure Pursuit controller \cite{4} was chosen. The Pure Pursuit controller is a popular choice for mobile robots due to its simplicity and effectiveness in tracking a reference path.

The Pure Pursuit controller is a type of proportional control that tries to steer a vehicle toward a reference path by adjusting the steering angle. The controller calculates the desired steering angle based on the current position of the vehicle and the desired point i.e. the lookahead point on the reference path. This point is calculated by finding the point on the path that is a certain distance ahead of the vehicle and aligns with the vehicle's current heading. This approach is simple and effective when the path is smooth and has gentle curves.

However, when the path contains sharp turns, the pure pursuit controller can deviate from the reference path, which can lead to collisions with static obstacles. This is because the controller assumes that the path is continuous and smooth, but in reality, sharp turns require the vehicle to slow down and turn more sharply to follow the path. If the controller continues to follow the reference path without adjusting for the sharp turn, the vehicle may overshoot the turn and collide with obstacles.

One way to mitigate this issue is to add safety measures. These can include adding obstacle detection and avoidance capabilities to the system, reducing the speed of the vehicle when sharp turns are detected, and adding a safety margin to the distance between the vehicle and obstacles. These measures can help prevent collisions and improve the safety of of the system.

Another pitfall is that too small a lookahead distance may cause the robot to oscillate or overshoot the path, while too large a lookahead distance may cause the robot to cut corners and deviate from the path. A possible solution to this is to set the lookahead distance based on the velocity of the robot and the curvature of the path. 

\subsection{Recomputation of Global Path}

Finally, while tracking the given global path using the said local planner, we keep checking whether there has been a change in the environment or whether the current path intersects any of the obstacles. If any of the conditions yields true, we generate the global path again.

\begin{figure}[H]
    \centering
    \includegraphics[width=0.7\linewidth]{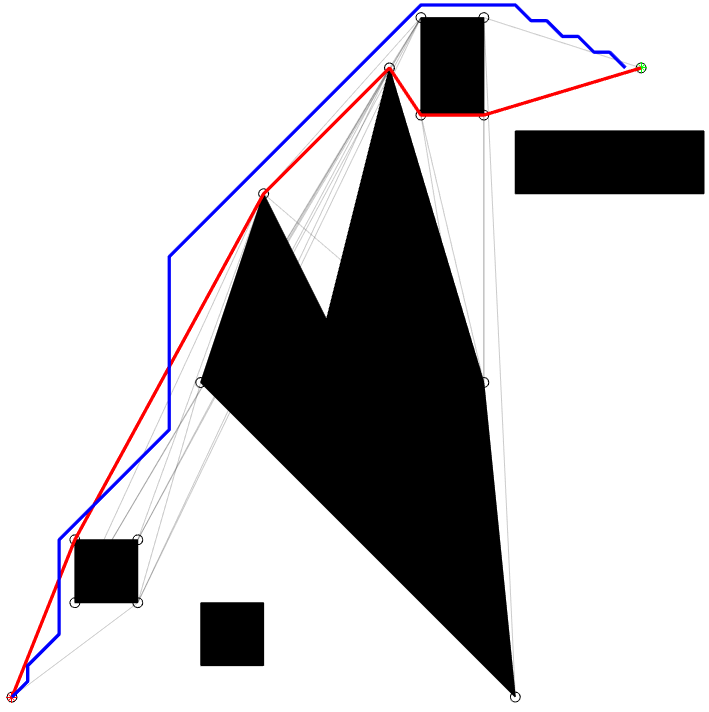}
    \caption{Figure comparing the global path generated by Minimal Construct algorithm (in red) and Grid-based A* algorithm (in blue)}
    \label{fig:results1}
\end{figure}

\begin{table}[H]
    \centering
    \begin{tabular}{ c c c }
	\hline
	& Minimal Construct & Grid based A*\\
	\hline
	Case 1		& $0.72, 0.70$	& $2.86, 2.46$\\
	Case 2		& $0.72, 0.38$	& $3.42, 1.72$\\
	Case 3		& $0.66, 0.37$	& $2.94, 1.98$\\
	Case 4		& $0.59, 0.27$	& $2.83, 1.40$\\
	\hline
    \end{tabular}
    \caption{Recomputation Time (sec) when Global Planner is called two times}
    \label{tab:recomputation}
\end{table}

\begin{figure}[H]
    \centering
    \includegraphics[width=0.8\linewidth]{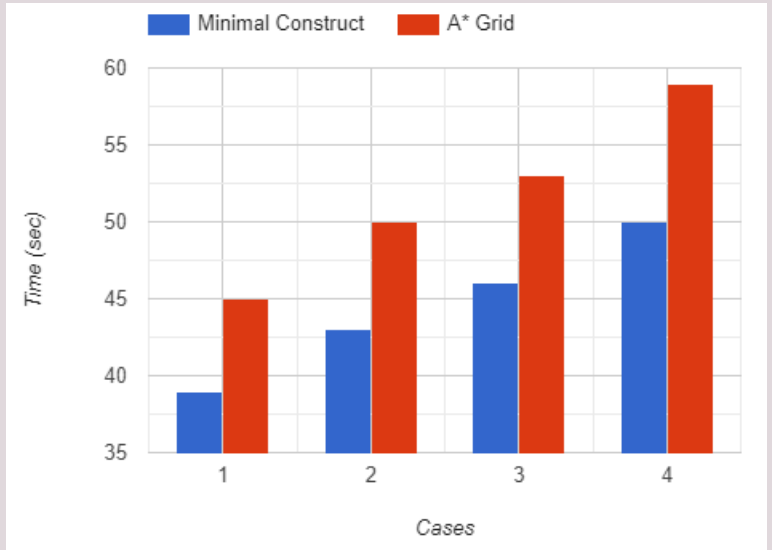}
    \caption{The figure shows the total time taken to reach the goal from the same start point for all the cases.}
    \label{fig:total_time}
\end{figure}

\section{Results}

The proposed approach was implemented in MATLAB and can be accessed through open-source code on \href{https://github.com/aditya-shirwatkar/CP230_HWS}{github}. The simulations were performed in two artificially generated maps of different natures i.e static and dynamic. 

In the case of Figure \ref{fig:results1} which is a static map, the path generated by the Minimal Construct is significantly shorter than the grid-based A*. And the graph explored is also minimal. Thereby motivating us to implement it in dynamic environments as well.

The simulations in the dynamically evolving map can be seen in Figure \ref{fig:results2}. It is assumed that the agent knows the changes in the environment. It can be seen that when the map changes the Grid based A* and Minimal Construct recompute the path. The time taken by Grid-based A* to recompute the new path is around 4 times more than the time taken by the Minimal Construct. A compilation of this recomputation time for different cases can be found in Table \ref{tab:recomputation}. This reduction in computation time will help in reducing the reaction time for the agent and also optimal utilization of onboard processing capacity. Also, the total time taken by the agent to reach a target in each case was plotted in figure \ref{fig:total_time}. Clearly, Minimal Construct outperforms the grid-based A*.


\section{Conclusion}

It was shown that the path search algorithm based on the Visibility Graph approach offers a significant reduction in computation time for finding the shortest path in polygonal maps. By adopting the philosophy of constructing only the required portion of the Visibility Graph, the Minimal Construct algorithm can efficiently calculate the shortest paths in cluttered and indoor environments. 

Polygonal planning and modeling provides an advantage of a non-discrete action set and a continuous representation of the world, resulting in smoother and optimal shortest paths with minimal path segments. Also preliminary results that leverage this polygonal representation where the map evolved dynamically were shown.

\section{Future Work}
The following extensions can be attempted as a part of future work.

$1)$ Incorporating a robust local planner like DWB \cite{6} or TEB-MPC \cite{7} \cite{8} for obstacle avoidance, as it can help improve the safety and efficiency of the system.

$2)$ Code-level optimizations in C++ can also help improve the performance of the system and reduce the computational load, which can be particularly important for real-time applications.

$3)$ Implementing the approach in different real-world scenarios can help evaluate the robustness and adaptability of the algorithm, and testing it with maps generated by state-of-the-art mapping algorithms can help validate its effectiveness in practical settings.

$4)$ Predicting potential collision points using mathematical expressions can also be a valuable addition to the system, especially for dynamically evolving maps \cite{9}. This can help ensure that the path planner can respond quickly to changes in the environment and avoid collisions with moving obstacles.

$5)$ Lastly, extending the algorithm to 3D polytopes can open up new applications for path planning, such as for UAVs. This can involve additional challenges such as dealing with complex terrain and varying altitude, but can also lead to exciting new possibilities for autonomous flight.

\bibliographystyle{IEEEtran}

\begin{thebibliography}{1}
\bibitem{1}
M. Quigley, B. Gerkey, K. Conley, J. Faust, T. Foote, J. Leibs, et al., "ROS: an open-source robot operating system", ICRA Workshop on Open Source Robotics, 2009.

\bibitem{2}
E. Marder-Eppstein, E. Berger, T. Foote, B. P. Gerkey and K. Konolige, "The office marathon: Robust navigation in an indoor office environment", Proc. of the IEEE Int. Conf on Robotics \& Automation (ICRA), 2010.

\bibitem{3}
Lozano-Pérez, Tomás, and Michael A. Wesley. "An algorithm for planning collision-free paths among polyhedral obstacles." Communications of the ACM 22, no. 10 (1979): 560-570.

\bibitem{4}
M. Missura, D. D. Lee and M. Bennewitz, "Minimal Construct: Efficient Shortest Path Finding for Mobile Robots in Polygonal Maps," 2018 IEEE/RSJ International Conference on Intelligent Robots and Systems (IROS), Madrid, Spain, 2018, pp. 7918-7923, doi: 10.1109/IROS.2018.8594124.

\bibitem{5}
R. Craig Coulter (1992). Implementation of the Pure Pursuit Path Tracking Algorithm [White paper]. Carnegie Mellon University.

\bibitem{6}
Fox, Dieter, Wolfram Burgard and Sebastian Thrun. “The dynamic window approach to collision avoidance.” IEEE Robotics Autom. Mag. 4 (1997): 23-33.

\bibitem{7}
Rösmann, Christoph, Frank Hoffmann, and Torsten Bertram. “Integrated Online Trajectory Planning and Optimization in Distinctive Topologies.” Robotics and Autonomous Systems. Elsevier BV, February 2017. doi:10.1016/j.robot.2016.11.007.

\bibitem{8}
C. Roesmann, W. Feiten, T. Woesch, F. Hoffmann, and T. Bertram, “Trajectory modification considering dynamic constraints of autonomous robots,” in ROBOTIK 2012; 7th German Conference on Robotics, pp. 1–6

\bibitem{9}
Thontepu, Phani, Bhavya Giri Goswami, Neelaksh Singh, Shyamsundar P I, Shyam Sundar M. G, Suresh Sundaram, Vaibhav Katewa, and Shishir Kolathaya. 2023. “Control Barrier Functions in UGVs for Kinematic Obstacle Avoidance: A Collision Cone Approach.” ArXiv:2209.11524 [Cs, Math], March. https://arxiv.org/abs/2209.11524.

\end{thebibliography}

\end{document}